\documentclass[journal]{IEEEtran}

\usepackage{amsmath,graphicx,algorithm,algpseudocode,amsfonts}
\usepackage{epstopdf}
\graphicspath{{Figures/}}
\usepackage[flushleft]{threeparttable}
\usepackage{multirow}
\usepackage{rotating}
\usepackage{caption}
\usepackage{subcaption}
\usepackage{tabularx}

\algnewcommand\algorithmicinput{\textbf{Input:}}
\algnewcommand\INPUT{\item[\algorithmicinput]}
\algnewcommand\algorithmicoutput{\textbf{Output:}}
\algnewcommand\OUTPUT{\item[\algorithmicoutput]}
\algnewcommand{\LineComment}[1]{\State \(\triangleright\) #1}

\begin{document}
% \title{Manifold to Manifold: A Geometry Aware Mapping Approach for Hypersepctral Image Classification}
\title{A Supervised Geometry-Aware Mapping Approach for Classification of Hyperspectral Images}


\author{Ramanarayan~Mohanty,~\IEEEmembership{Student Member,~IEEE,}
        S~L~Happy,~\IEEEmembership{Student Member,~IEEE,}
        and~Aurobinda~Routray,~\IEEEmembership{Member,~IEEE}% <-this % stops a space
\thanks{Ramanarayan Mohanty was with the Advanced Technology Development Centre, Indian Institute of Technology Kharagpur,
West Bengal, 721302 India e-mail: ramanarayan@iitkgp.ac.in.}}% <-this % stops a space

\markboth{Journal of \LaTeX\ Class Files,~Vol.~14, No.~8, August~2015}%
{Shell \MakeLowercase{\textit{et al.}}: Bare Demo of IEEEtran.cls for IEEE Journals}

\maketitle

\begin{abstract}
The lack of proper class discrimination among the Hyperspectral (HS) data points poses a potential challenge in HS classification. To address this issue, this paper proposes an optimal geometry-aware transformation for enhancing the classification accuracy. The underlying idea of this method is to obtain a linear projection matrix by solving a nonlinear objective function based on the intrinsic geometrical structure of the data. The objective function is constructed to quantify the discrimination between the points from dissimilar classes on the projected data space. Then the obtained projection matrix is used to linearly map the data to more discriminative space. The effectiveness of the proposed transformation is illustrated with three benchmark real-world HS data sets. The experiments reveal that the classification and dimensionality reduction methods on the projected discriminative space outperform their counterpart in the original space.

\end{abstract}
\begin{IEEEkeywords}
	Dimensionality reduction, geometry-aware mapping, hyperspectral classification, manifold, nonlinear objective function.
\end{IEEEkeywords}

\IEEEpeerreviewmaketitle

\section{Introduction}	\label{sec:intro}

The multi-path scattering of light within a pixel \cite{keshava2002spectral}, bidirectional reflectance distribution \cite{goodin2004effect}, and the heterogeneity of sub-pixel constituents \cite{bachmann2005exploiting} are the major concerns in the hyperspectral (HS) data classification. These nonlinearity properties naturally place the HS data on a non-euclidean space.
%The major limiting factor in the classification task of Hyperspectral (HS) data is its nonlinear characteristics, which occurs due to various sources like the multi-path scattering of light within a pixel \cite{keshava2002spectral}, bidirectional reflectance distribution \cite{goodin2004effect} and the heterogeneity of sub-pixel constituents \cite{bachmann2005exploiting}. These nonlinearity properties naturally place the HS data on a non-euclidean space. %Despite the nonlinearity property, the HS data are typically of high dimensional and redundant. 
Handling these high dimensional redundant data in a non-euclidean space is one of the major bottlenecks in HS data analysis. 
%or on Riemannian manifold (e.g., subspaces form Grassmannian). Despite the nonlinearity property the HS data are typically high dimensional and redundant. Handling these high dimensional redundant data on Riemannian manifold is one of the major bottleneck in HS data analysis.

%\textbf{Existing methods in linear DR and non linear DR for non euclidean spaces}

Typically, HS classification consists of dimensionality reduction (DR) and subsequent classification operation. %The DR methods are useful in HS data analysis to address the problem of the curse of dimensionality. 
The popular DR methods such as principal component analysis (PCA) \cite{martinez2001pca} and linear discriminant analysis (LDA) \cite{belhumeur1997eigenfaces} are linear and operate on Euclidean structures. These linear DR methods skip the curved nonlinear structures of the HS data. On the other hand, manifold learning helps in recovering compact, meaningful low dimensional structures from those complex high dimensional data from a non-euclidean space. The manifold learning methods consider the real world high dimensional data to be generated with a few degrees of freedom \cite{zheng2009statistical}. This leads to the projection of the data into lower dimensional space while preserving their underlying geometrical structure \cite{lin2008riemannian}. 

%\textbf{Non linear methods consider local and global geometry approach and problem with these approaches}

Several state-of-art techniques for DR use manifold learning, such as local linear embedding (LLE) \cite{han2005nonlinear}, isometric feature mapping (ISOMAP) \cite{tenenbaum2000global}, Laplacian eigenmap (LE) \cite{qian2007new}, local tangent space alignment (LTSA) \cite{zhang2004principal}, local scaling cut (LSC) and semi-supervised local scaling cut (SSLSC) \cite{zhang2013semisupervised} etc. These algorithms assume that the input data lie on or close to a smooth low dimensional manifold \cite{lin2008riemannian} \cite{ma2010local}. %\textit{\textbf{Cite the papers related to HSI, if there is any.}}
Some of these algorithms use local approaches like LE, LLE, LTSA, LSC, SSLSC etc., while others use the global approaches such as ISOMAP \cite{ma2015local}. These local approaches use spectral embedding method to retain the local geometry of the data while projecting them to lower dimensions. The embedding task, in these algorithms, is reduced to the form of an eigen decomposition problem under unit covariance constraint \cite{yan2007graph}. This imposed covariance constraint lose the aspect ratio of the embedding data and the underlying manifold can not reflect its original global shape. 
On the other hand, the global approach like ISOMAP preserves the metrics at all scales and give a better embedding. However, ISOMAP is only applied to intrinsically flat structures (cylinders, cones, etc) \cite{yan2007graph}. Therefore, the majority of the manifold based DR algorithms give undesired results in HS data classification due to the above-said issues. In order to address this, mapping the data from existing space to more discriminative space appears to be a promising direction. 
%\textbf{Q:What is your method using (local/global)? Explain why? How is it improving on others/ reasons?
%}

%\textbf{Main problem in existing manifold based DR approaches}

%\textbf{This paragraph is wrong. I believe the manifold based approaches will work better than linear projection methods.  Instead write about the computational complexities in manifold approach, its disadvantages. you can compare the current results with [16] to show our advantages even with linear mapping (in result section), if possible.
%Additionally, write that the nonlinear manifold structure is not proper with less training data points. However, linear projection is okay with small sample size. High dimension has disadvantages in distance measurement, which we need for manifold construction.}
A majority of the manifold learning (ML) algorithms first flatten the non-euclidean structure by different methods such as tangent space computation and Hilbert space embedding \cite{harandi2014manifold}. The flattened non-euclidean structure then projected to a lower dimensional more discriminative space or directly apply the classification algorithms on that flattened structure. This flattening process mainly incurs distortions in the data geometry on the non-euclidean space \cite{harandi2014manifold}. Then this distortion imparts a negative impact on the projection operation as well as the classification which gives undesired results. Additionally, the nonlinear manifold structure is also not properly constructed with less training data points with disruptive geometry in their original data space. %However, linear projection is okay with small sample size. High dimension has disadvantages in distance measurement, which we need for manifold construction.

%\textbf{Proposed approach}

In this work, we propose a geometry-aware mapping approach that maps the data from the original data space to a new more discriminative space to achieve better performance in subsequent DR and classification tasks.
The mapping procedure preserves the original geometry of the data to a great extent in the discriminative space due to the linear projection. However, we employ the nonlinear cost function based on the local property by embedding a weighted graph on the data for computing the projection matrix. Moreover, the positioning of data from similar and dissimilar classes in the local neighborhood is also incorporated in the cost function making it more reliable for obtaining optimal discriminative directions. Here the class labels are used to build an affinity function to encode the inter-class and intra-class similarity. The projection matrix is obtained using these similarity matrices without disrupting the underlying geometry.
%distorting the orientation of the data geometry. 
Thus, the projected data lie on a more discriminative space improve the subsequent data interpretation.

\section{Proposed Geometry Aware Mapping Technique}		\label{sec:proposed_method}
The DR methods aim at finding a projection matrix to project the data points to a new space with less degree of freedoms. However, some information is lost in this process, which in turn affects the performance of the system. To address this issue, we aim to project the data to a new space with reduced dimension that will minimize the information loss and maximize the class discrimination by preserving the geometry of the data points. Thus, our goal is to obtain a projection matrix that will improve the classification performance in the projected space without distorting the data geometry. 

Given a set of data points $X= \{x_1, x_2, ...,x_p\}; x_i \in R^{n}$ with class labels $y_i \in \{1,2,...,c\}$, our objective is to find the mapping matrix $U$, such that, the resulting space enhances some interesting structure of the original data with minimum information loss and improves the classification performance. 
Thus, we seek to learn the mapping matrix $U \in R^{n \times m}$, of a generic mapping $f: S^{n} \times R^{n \times m} \rightarrow  S^{m}$, $m < n$ and is defined as 
\begin{equation}
f(X,U) = U^{T}X
\end{equation}
%Since $U$ is a linear projection matrix in $R^{n \times m}$, the distance between the points in the projected space will remain the same. 
To improve the classification accuracy, we need to determine the directions along which different class points can be discriminated easily. Class overlapping in HS data is a big issue because of the complex structure of the data. In such cases, we can focus on directions which will improve the discrimination in closely situated points from dissimilar classes. Thus, the intrinsic nonlinear geometrical structure of the data can be utilized to improve class discrimination locally.

An unsupervised way of achieving DR would consider the following cost function,
\begin{equation} \label{eq:unsuper-costFun}
\min_U L(U) = \sum_{i,j}||{U^T}{x_i}-{U^T}{x_j}||_{F}^2
\end{equation}
However, with the knowledge of the class labels of the training data, we can formulate a non-linear cost function to achieve the optimal class discrimination in the projected space.
To this end, we use the class labels along with the affinity of the data points to formulate the cost function. 

We encode the underlying geometric structure of the data by an undirected graph reflected by the affinity matrix $A \in R^{n \times n}$. The element $A_{ij}$ of affinity matrix $A$ measures the affinity between data point $x_i$ and $x_j$. In particular, we use the class labels to compute the affinity function through intra-class $(g_w)$ and inter-class $(g_b)$ similarities \cite{weinberger2009distance} \cite{weinberger2006distance}. The binary functions $g_w$ and $g_b$ are computed as 
\begin{equation}
\begin{split}
g_{w}(x_i, x_j) = \left\{ {\begin{array}{*{20}{c}}
	{1,\,\,\,\,\, \mbox{if}\,\,{x_i}\, \in \,{N_w}({x_j})\,\mbox{or}\,\,\,{x_j}\, \in \,{N_w}({x_i})}\\
	{0,\,\,\,\,\,\,\,\,\,\,\,\,\,\,\,\,\,\,\,\,\,\,\,\,\,\,\,\,\,\,Otherwise\,\,\,\,\,\,\,\,\,\,\,\,\,\,\,\,\,\,\,\,\,\,\,\,\,\,\,}
	\end{array}} \right. \\
g_{b}(x_i, x_j) = \left\{ {\begin{array}{*{20}{c}}
	{1,\,\,\,\,\, \mbox{if}\,\,{x_i}\, \in \,{N_b}({x_j})\,\mbox{or}\,\,\,{x_j}\, \in \,{N_b}({x_i})}\\
	{0,\,\,\,\,\,\,\,\,\,\,\,\,\,\,\,\,\,\,\,\,\,\,\,\,\,\,\,\,\,\,Otherwise\,\,\,\,\,\,\,\,\,\,\,\,\,\,\,\,\,\,\,\,\,\,\,\,\,\,\,}
	\end{array}} \right.
\end{split}
\end{equation}
where $N_w(x_i)$ is the set of $v_w$ nearest neighbors of $x_i$ that belong to the same class as of $x_i$, $N_b(x_i)$ is the set of $v_b$ nearest neighbors of $x_i$ that belong to different classes, other than the class of $x_i$. %The $L2$-norm is used to compute the nearest neighbors. 
The affinity function is then defined as
\begin{equation}
A_{ij} = g_{w}(x_i, x_j) - g_{b}(x_i, x_j)
\end{equation}

After obtaining $A$, we propose to modify the cost function in (\ref{eq:unsuper-costFun}) by incorporating the supervised local properties for computing the projection matrix. Thus, we formulate the cost function as %search for an embedding, such that the affinity between the pair of classes is reflected by a measure of similarity on the new projected space. This lets us to write the loss function of the form 
\begin{equation}
L(U) = \sum_{i,j} {A_{ij}}||{U^T}{x_i}-{U^T}{x_j}||_{F}^2
\end{equation}
%\textbf{To avoid degeneracies in the resulting embedding we need to ensure that $U^TX > 0$, $\forall X \in S^n$ (what do you mean?). Hence $U$ is needed to be a full rank matrix.} 
An orthonormal constraint is imposed on the objective function $L(U)$ i.e. $U^TU = I_m$. Hence the final projection matrix can be obtained by minimizing the $L(U)$ w.r.t $U$ such that $U^TU$=$I_m$. That is 
\begin{equation} \label{eq:opti_prblm}
U = \begin{array}{*{20}{c}}
{\min }\\
{U \in {R^{n \times m}}}
\end{array}L(U)\,\,\,\,\,\, \mbox{s.t.}\,\,\,\,{U^T}U = {I_m}\,
\end{equation}
% \begin{equation} \label{eq:opti_prblm}
% \begin{split}
% \begin{array}{*{20}{c}}
% {\min }\\
% {U \in {R^{n \times m}}}
% \end{array}L(U) \\
% \mbox{s.t.} \,\, U^TU=I_m
% \end{split}
% \end{equation}
Here in (\ref{eq:opti_prblm}), $L(U)$ is the optimization function with unitary constraint. Here the objective function forces the distance between the geometrically close points of the same class to be closer, while increasing the distance between closely situated dissimilar class points.
The addition of nonlinear weighted graph $A_{ij}$ with the linear cost function of (\ref{eq:unsuper-costFun}) make the proposed cost function  nonlinear in nature. Moreover, the complexity of this cost function $L(U)$ in (\ref{eq:opti_prblm}) leads us to solve this using nonlinear optimization technique. 

%\textbf{The cost function with unitary constraint is either formulated as an optimization problem on Stiefel  or Grassmann space \cite{edelman1998geometry}, \cite{absil2009optimization}. Since the cost function $L(U)$ uses the Frobenius norm and it possesses rotational invariant property i.e $L(U) = L(UR)$ for $R^{T}R = RR^{T} = I_m$, then this makes the problem in Grassmann space. Normally a Grassmannian $\mathcal{G}(m,n)$ is the space of $m$ dimensional subspaces of $R^n$ for $0 < m \le n$ \cite{absil2009optimization}. 
Taking inspiration from \cite{harandi2014manifold} and \cite{harandi2017dimensionality}, here we use a Conjugate Gradient (CG) method to solve the optimization problem (\ref{eq:opti_prblm}). In this work, we use \textit{manopt} optimization toolbox \cite{boumal2014manopt} to implement the CG method. The CG method completely relies on the gradient on the nonlinear space. To this end, the gradient of a function is computed as  
\begin{equation} \label{eq:gradient}
grad~L(U) = (I_n - UU^T) \nabla_U(L)
\end{equation}
where $\nabla_U(L)$ is the usual Jacobian matrix at $U$ and the Jacobian matrix for Frobenius norm is computed as 
\begin{equation}\label{eq:jacobian_mat}
\nabla_U(L)= 2U^T(x_i-x_j)(x_i-x_j)'  
\end{equation}

%#########################################################################
% \begin{figure}[!h]
% 	\centering
% 	\begin{subfigure}{.24\textwidth}
% 		\centering
% 		\includegraphics[width=.98\linewidth]{Figures/Orig_Indp_tSNE_Cls_4_13_dot_crpd.png}
% 		\caption{Original}%{Botswana dataset}
% 		\label{fig:IndP10DR}
% 	\end{subfigure}%
% 	\begin{subfigure}{.24\textwidth}
% 		\centering
% 		\includegraphics[width=.98\linewidth]{Figures/Mapped_Indp_tSNE_Cls_4_13_dot_crpd.png}
% 		\caption{Mapped}%Indian-pines dataset}
% 		\label{fig:Bow10DR}
% 	\end{subfigure}
%     \caption{Variations in orientation of data before mapping and after mapping} 
% 	\label{fig:orig_mapped}
% \end{figure}
%#########################################################################################################

Finally, we project the HS data from its original space into a new discriminative space spanned by $U$ by retaining their geometrical structure to achieve better discriminative property for subsequent DR and classification operation. 
In our experiments, we found that the optimization cost is minimum when the reduced dimension in the proposed method is one less than the dimension of the data. This can be simply explained as the dimension in which the information loss will be minimum, while improving the class discrimination. Therefore, similar setting is adopted in all our experiments.

\section{Experiments}
\label{sec:result}
%\input{result}

% \usepackage{graphicx}
\begin{table}[htp]
	\centering
	\caption{The effect of variation of the number of nearest neighbor ($N_w$ and $N_b$) points on overall accuracy (OA)}
	\label{tab:knn_OA}
	\resizebox{0.48\textwidth}{!}{%
		\begin{tabular}{c|cc|cc|cc}
			\hline
			Datasets & \multicolumn{2}{c|}{Indian Pines} & \multicolumn{2}{c|}{Botswana} & \multicolumn{2}{c}{Pavia University} \\ \hline
			$N_w = N_b$ & \begin{tabular}[c]{@{}c@{}}Original\\ Space\end{tabular} & \begin{tabular}[c]{@{}c@{}}Projected\\ Space\end{tabular} & \begin{tabular}[c]{@{}c@{}}Original\\ Space\end{tabular} & \begin{tabular}[c]{@{}c@{}}Projected\\ Space\end{tabular} & \begin{tabular}[c]{@{}c@{}}Original\\ Space\end{tabular} & \begin{tabular}[c]{@{}c@{}}Projected\\ Space\end{tabular} \\ \hline
			3 & 52.86 & 60.11 & 85.96 & 88.01 & 68.90 & 72.27 \\ %\hline
			5 & 53.50 & 57.82 & 87.65 & 86.91 & 66.30 & 68.82 \\ %\hline
			7 & 52.75 & 58.57 & 86.44 & 86.88 & 65.98 & 74.32 \\ %\hline
			9 & 56.23 & \textbf{60.31} & 89.10 & \textbf{90.21} & 67.30 & 73.93 \\ %\hline
			11 & 54.87 & 57.53 & 86.06 & 88.18 & 67.27 & 73.19 \\ %\hline
			13 & 55.94 & 59.44 & 86.70 & 86.14 & 67.08 & \textbf{74.05} \\ \hline
		\end{tabular}%
	}
\end{table}

%#########################################################################################################

%where as for the KNN classifier the K value is set for three instances $1, 3$ and $5$. Similarly the discriminant classifier    

% \usepackage{multirow}
\begin{table*}[htp]
	\centering
	\caption{Experimental results (AA:average accuracy, $\kappa$:kappa, and OA:overall accuracy) with different classifiers}
	\label{tab:Accuracy_comparison}
	\scalebox{0.83}{
		\begin{tabular}{ccccccccccccccccccc}
			\hline
			\multicolumn{1}{|p{0.2cm}|}{\multirow{4}{*}{\parbox{0.16cm}{\centering Classi \\ fiers}}} & \multicolumn{18}{c|}{Datasets}                                                                                                                                                                                                      \\ \cline{2-19} 
			\multicolumn{1}{|c|}{}                            & \multicolumn{6}{c|}{Indian Pines}                                                                                                 & \multicolumn{6}{|c|}{Botswana}                                                                                                     & \multicolumn{6}{|c|}{Pavia University}                                                                                             \\ \cline{2-19}
			\multicolumn{1}{|l|}{}                             & \multicolumn{3}{c|}{Original Space} & \multicolumn{3}{|c|}{Projected Space} & \multicolumn{3}{|c|}{Original Space} & \multicolumn{3}{|c|}{Projected Space} & \multicolumn{3}{|c|}{Original Space} & \multicolumn{3}{|c|}{Projected Space} \\ \cline{2-19} 
			\multicolumn{1}{|c|}{}                            & \multicolumn{1}{c}{AA} & \multicolumn{1}{c}{OA}   & \multicolumn{1}{p{0.18cm}|}{$\kappa$} & \multicolumn{1}{|c}{AA} & \multicolumn{1}{c}{OA}   & \multicolumn{1}{p{0.18cm}|}{$\kappa$} & \multicolumn{1}{|c}{AA} & \multicolumn{1}{c}{OA}   & \multicolumn{1}{p{0.18cm}|}{$\kappa$} & \multicolumn{1}{|c}{AA}   & \multicolumn{1}{p{0.18cm}}{OA} & \multicolumn{1}{c|}{$\kappa$} & \multicolumn{1}{|p{0.18cm}}{AA} & \multicolumn{1}{c}{OA}   & \multicolumn{1}{p{0.18cm}|}{$\kappa$} & \multicolumn{1}{|p{0.18cm}}{AA} & \multicolumn{1}{c}{OA}   & \multicolumn{1}{p{0.18cm}|}{$\kappa$}  \\ \hline
			
			\multicolumn{1}{|c|}{SVM}                                                & \multicolumn{1}{|c}{ 69.16}      & \multicolumn{1}{c}{56.43}     & \multicolumn{1}{c|}{0.5123}     & \multicolumn{1}{|c}{70.35}      & \multicolumn{1}{c}{60.53}      & \multicolumn{1}{c|}{0.5629}     & \multicolumn{1}{|c}{89.18}      & \multicolumn{1}{c}{89.12}     & \multicolumn{1}{c|}{0.8904}     & \multicolumn{1}{|c}{90.58}      & \multicolumn{1}{c}{90.39}      & \multicolumn{1}{c|}{0.8956}     & \multicolumn{1}{|c}{74.47}      & \multicolumn{1}{c}{ 66.45}     & \multicolumn{1}{c|}{0.5813}     & \multicolumn{1}{|c}{74.61}      & \multicolumn{1}{c}{70.17}      & \multicolumn{1}{c|}{0.6373}     \\ \hline
			\multicolumn{1}{|c|}{1NN}                                                & \multicolumn{1}{|c}{66.93}      & \multicolumn{1}{c}{53.51}     & \multicolumn{1}{c|}{0.4817}     & \multicolumn{1}{|c}{\textbf{75.87}}      & \multicolumn{1}{c}{\textbf{77.66}}      & \multicolumn{1}{c|}{\textbf{0.7428}}     & \multicolumn{1}{|c}{89.08}      & \multicolumn{1}{c}{87.49}     & \multicolumn{1}{c|}{0.8642}     & \multicolumn{1}{|c}{\textbf{96.67}}      & \multicolumn{1}{c}{\textbf{96.45}}      & \multicolumn{1}{c|}{\textbf{0.9614}}     & \multicolumn{1}{|c}{74.38}      & \multicolumn{1}{c}{62.85}     & \multicolumn{1}{c|}{0.5436}     & \multicolumn{1}{|c}{\textbf{90.03}}      & \multicolumn{1}{c}{\textbf{89.12}}      & \multicolumn{1}{c|}{\textbf{0.8684}}     \\ \hline
			\multicolumn{1}{|c|}{3NN}                                                & \multicolumn{1}{|c}{63.31}      & \multicolumn{1}{c}{50.30}     & \multicolumn{1}{c|}{0.4455}     & \multicolumn{1}{|c}{64.16}      & \multicolumn{1}{c}{63.79}      & \multicolumn{1}{c|}{0.5996}     & \multicolumn{1}{|c}{88.52}      & \multicolumn{1}{c}{86.88}     & \multicolumn{1}{c|}{0.8577}     & \multicolumn{1}{|c}{92.57}      & \multicolumn{1}{c}{91.86}      & \multicolumn{1}{c|}{0.9116}     & \multicolumn{1}{|c}{72.16}      & \multicolumn{1}{c}{57.80}     & \multicolumn{1}{c|}{0.4906}     & \multicolumn{1}{|c}{82.78}      & \multicolumn{1}{c}{77.94}      & \multicolumn{1}{c|}{0.7350}     \\ \hline
			\multicolumn{1}{|c|}{5NN}                                                & \multicolumn{1}{|c}{57.24}      & \multicolumn{1}{c}{43.67}     & \multicolumn{1}{c|}{0.3736}     & \multicolumn{1}{|c}{58.89}      & \multicolumn{1}{c}{51.70}      & \multicolumn{1}{c|}{0.4683}     & \multicolumn{1}{|c}{87.20}      & \multicolumn{1}{c}{85.51}     & \multicolumn{1}{c|}{0.8428}     & \multicolumn{1}{|c}{90.37}      & \multicolumn{1}{c}{89.54}      & \multicolumn{1}{c|}{0.8865}     & \multicolumn{1}{|c}{70.60}      & \multicolumn{1}{c}{55.61}     & \multicolumn{1}{c|}{0.4676}     & \multicolumn{1}{|c}{78.91}      & \multicolumn{1}{c}{72.35}      & \multicolumn{1}{c|}{0.6694}     \\ \hline
			\multicolumn{1}{|c|}{LDC}                                                & \multicolumn{1}{|c}{52.63}      & \multicolumn{1}{c}{43.40}     & \multicolumn{1}{c|}{0.3676}     & \multicolumn{1}{|c}{55.63}      & \multicolumn{1}{c}{44.78}      & \multicolumn{1}{c|}{0.3907}     & \multicolumn{1}{|c}{78.67}      & \multicolumn{1}{c}{75.97}     & \multicolumn{1}{c|}{0.7399}     & \multicolumn{1}{|c}{80.04}      & \multicolumn{1}{c}{77.80}      & \multicolumn{1}{c|}{0.7597}     & \multicolumn{1}{|c}{67.39}      & \multicolumn{1}{c}{56.67}     & \multicolumn{1}{c|}{0.4740}     & \multicolumn{1}{|c}{72.23}      & \multicolumn{1}{c}{65.76}      & \multicolumn{1}{c|}{0.5894}     \\ \hline
			\multicolumn{1}{|c|}{QDC}                                                & \multicolumn{1}{|c}{53.96}      & \multicolumn{1}{c}{44.05}     & \multicolumn{1}{c|}{0.3736}     & \multicolumn{1}{|c}{58.14}      & \multicolumn{1}{c}{45.25}      & \multicolumn{1}{c|}{0.3966}     & \multicolumn{1}{|c}{81.10}      & \multicolumn{1}{c}{78.99}     & \multicolumn{1}{c|}{0.7722}     & \multicolumn{1}{|c}{82.38}      & \multicolumn{1}{c}{80.70}      & \multicolumn{1}{c|}{0.7909}     & \multicolumn{1}{|c}{70.67}      & \multicolumn{1}{c}{61.66}     & \multicolumn{1}{c|}{0.5251}     & \multicolumn{1}{|c}{74.12}      & \multicolumn{1}{c}{67.85}      & \multicolumn{1}{c|}{0.6114}     \\ \hline
			\multicolumn{1}{|c|}{TREE}                                               & \multicolumn{1}{|c}{50.36}      & \multicolumn{1}{c}{39.64}     & \multicolumn{1}{c|}{0.3364}     & \multicolumn{1}{|c}{54.98}      & \multicolumn{1}{c}{43.62}      & \multicolumn{1}{c|}{0.3797}     & \multicolumn{1}{|c}{80.60}      & \multicolumn{1}{c}{79.40}     & \multicolumn{1}{c|}{0.7763}     & \multicolumn{1}{|c}{81.06}      & \multicolumn{1}{c}{80.41}      & \multicolumn{1}{c|}{0.7873}     & \multicolumn{1}{|c}{61.05}      & \multicolumn{1}{c}{51.88}     & \multicolumn{1}{c|}{0.4152}     & \multicolumn{1}{|c}{63.47}      & \multicolumn{1}{c}{57.30}      & \multicolumn{1}{c|}{0.4886}     \\ \hline
		\end{tabular}}
	\end{table*}
	
	%% ############################################################################################################################################################################################################################
	%#########################################################################
	\begin{figure*}[!h]
		\centering
		\begin{subfigure}{.33\textwidth}
			\centering
			\includegraphics[width=.98\linewidth]{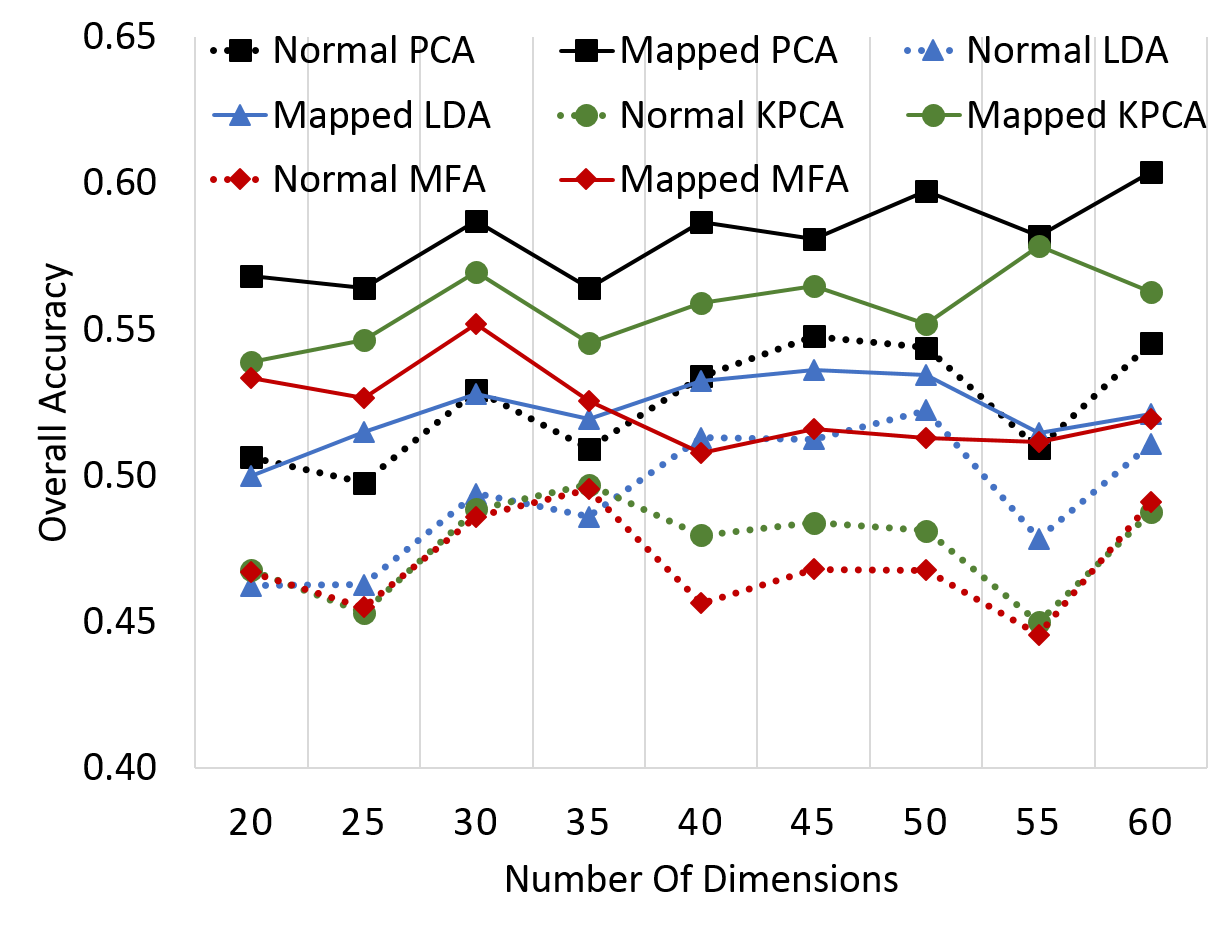}
			\caption{}%{Botswana dataset}
			\label{fig:IndP10DR}
		\end{subfigure}%
		\begin{subfigure}{.33\textwidth}
			\centering
			\includegraphics[width=.98\linewidth]{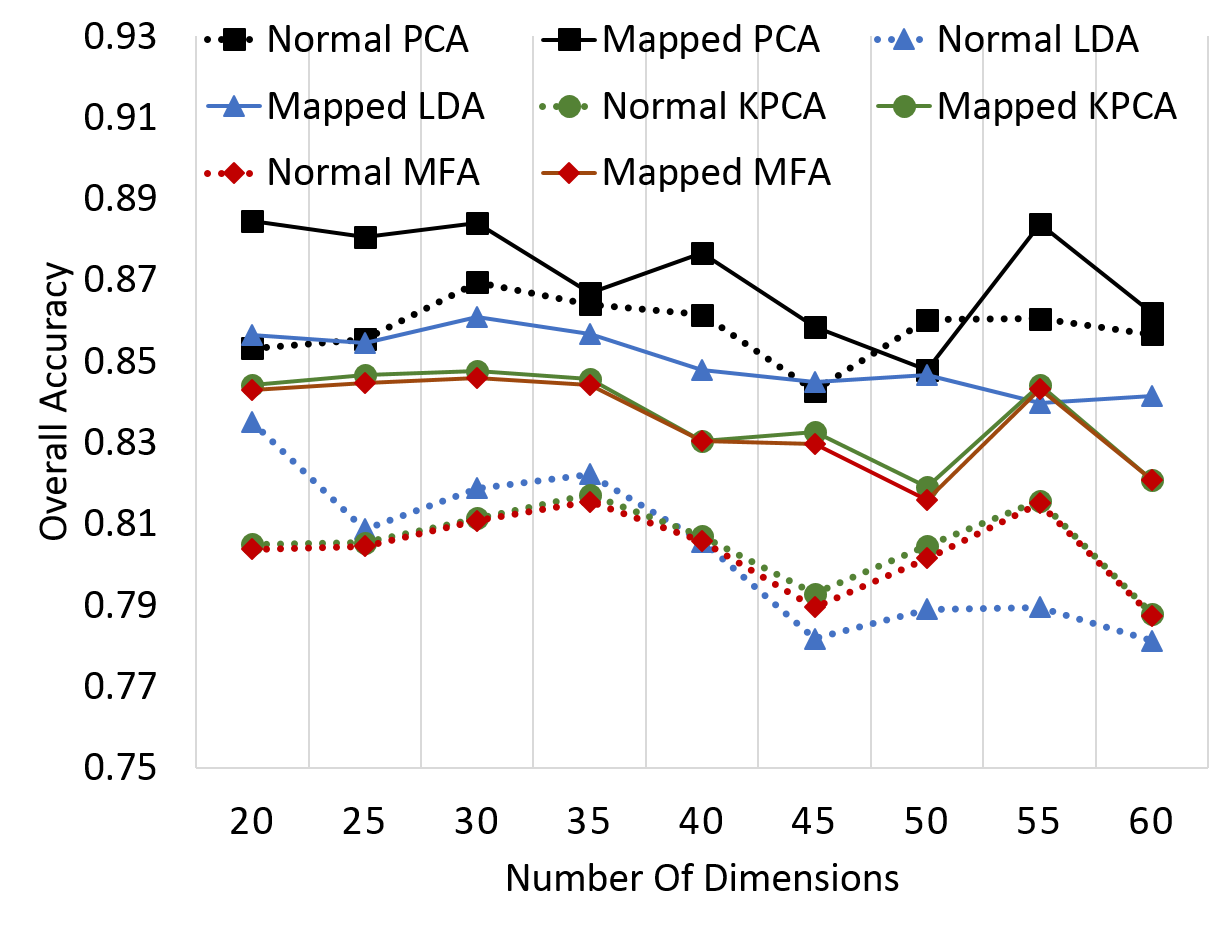}
			\caption{}%Indian-pines dataset}
			\label{fig:Bow10DR}
		\end{subfigure}
		\begin{subfigure}{.33\textwidth}
			\centering
			\includegraphics[width=.98\linewidth]{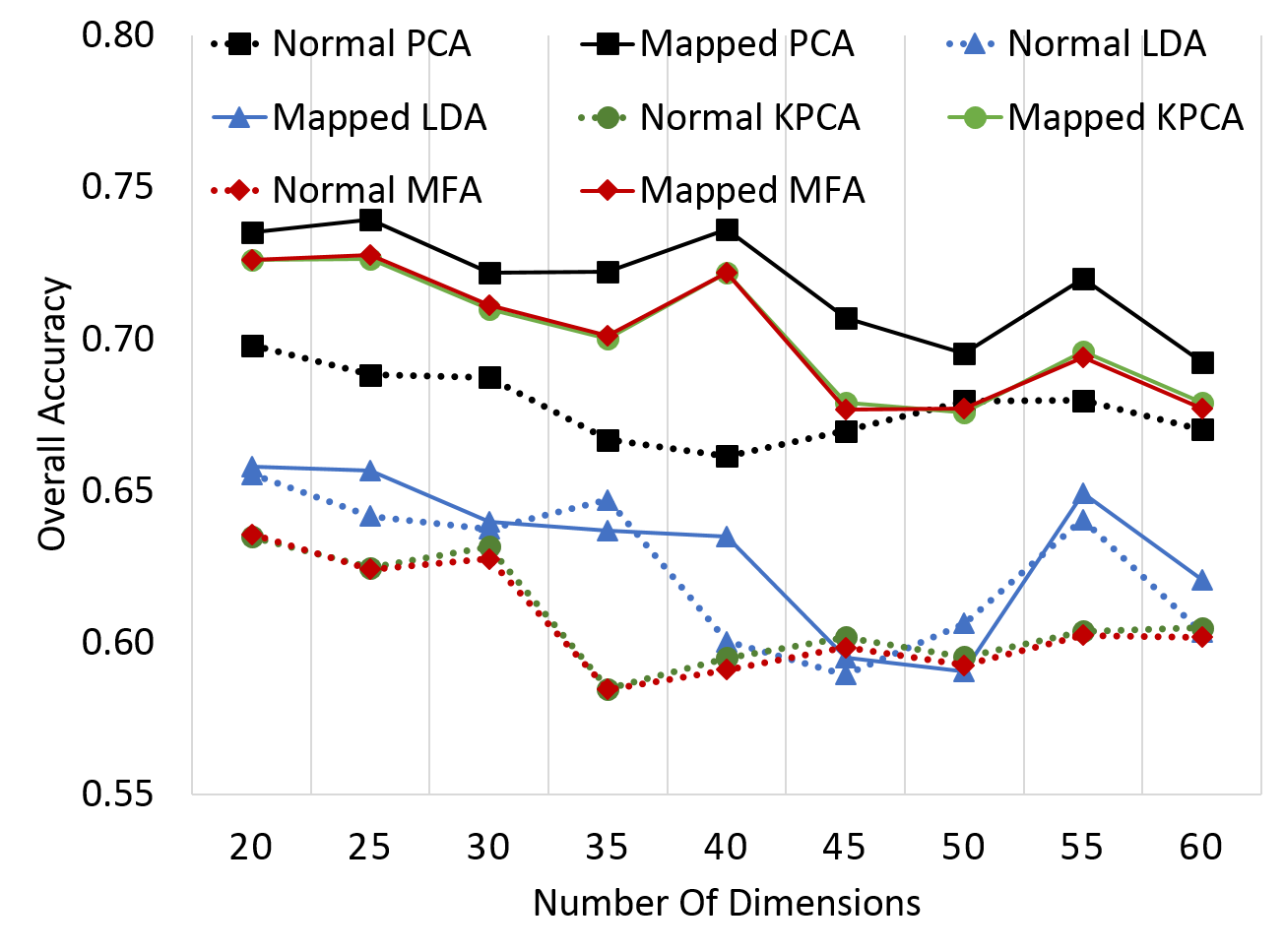}
			\caption{}%Pavia University dataset}
			\label{fig:PavU10DR}
		\end{subfigure}
		\caption{Comparison of classification performances with four different DR techniques (PCA, LDA, KPCA and MFA) w.r.t various dimensions on three different HS datasets (Indian Pines (\ref{fig:IndP10DR}), Botswana (\ref{fig:Bow10DR}) and Pavia University (\ref{fig:PavU10DR})) .(Notations:- e.g. Normal PCA: PCA in original space (dotted line), Mapped PCA: PCA in projected discriminative space (solid line))} 
		\label{fig:DRhsi10}
	\end{figure*}
	%#########################################################################################################
	
	\subsection{Dataset Description}
	Three HS image datasets were used in our experiments, such as Indian Pine, Botswana, and Pavia university\footnote{http://www.ehu.eus/ccwintco/index.php?title=Hyperspectral\_Remote\\\_Sensing\_Scenes}. The size of the Indian Pine dataset is $145 \times 145$ pixels in the spatial domain and $200$ bands in spectral domain with $16$ different classes. Botswana dataset consists of $ 1476 \times 256$ pixels in the spatial domain and $145$  bands in spectral domain with $14$ different classes. Similarly, Pavia university consists of $610 \times 610$ spatial pixels with $103$ spectral bands of $9$ different classes. 
	%#####################################################################################################
	% \textbf{What parameter used for finding $N_b,N_w$.
	% }
	% \begin{table}[h!]
	% \centering
	% \caption{Computation time of the proposed mapping approach for two data samples for three different datasets}
	% \label{tab:time_dataSample}
	% \begin{tabular}{cclclcl}
	% \hline
	% Data samples & \multicolumn{2}{|c}{Indian Pines} & \multicolumn{2}{|c|}{Botswana} & \multicolumn{2}{c}{Pavia University} \\ \hline
	% 10 & \multicolumn{2}{|c}{327.51} & \multicolumn{2}{|c|}{189.30} & \multicolumn{2}{c}{70.29} \\ \hline
	% 15 & \multicolumn{2}{|c}{485.70} & \multicolumn{2}{|c|}{290.83} & \multicolumn{2}{c}{86.22} \\ \hline
	% \end{tabular}
	% \end{table}
	%####################################################################################################
	%$$$$$$$$$$$$$$$$$$$$$$$$$$$$$$$$$$$$$$$$$$$$$$$$$$$$$$$$$$$$$$$$$$$$$$$$$$$$$$$$$$$$$$$$$$$$$$$$$$$
	\begin{table}[!h]
		\centering
		\caption{Computation time of the proposed mapping approach (for training and testing) for different number of training data samples}
		\label{tab:time}
		\begin{tabular}{l|c|c|c|c|c}
			\hline
			No of Train Samples & 5 & 8 & 10 & 15 & 20 \\ \hline
			Training Time (Sec) & 312.69 & 509.27 & 558.66 & 918.28 & 1311.6 \\ %\hline
			Testing Time (Sec) & 0.0802 & 0.0852 & 0.0819 & 0.0854 & 0.0928 \\ \hline
		\end{tabular}
	\end{table}
	%$$$$$$$$$$$$$$$$$$$$$$$$$$$$$$$$$$$$$$$$$$$$$$$$$$$$$$$$$$$$$$$$$$$$$$$$$$$$$$$$$$$$$$$$$$$$$$$$$$$

	%#########################################################################
	\begin{figure}[!h]
		\centering
		%\begin{subfigure}{.45\textwidth}
		\centering
		\includegraphics[width=.98\linewidth]{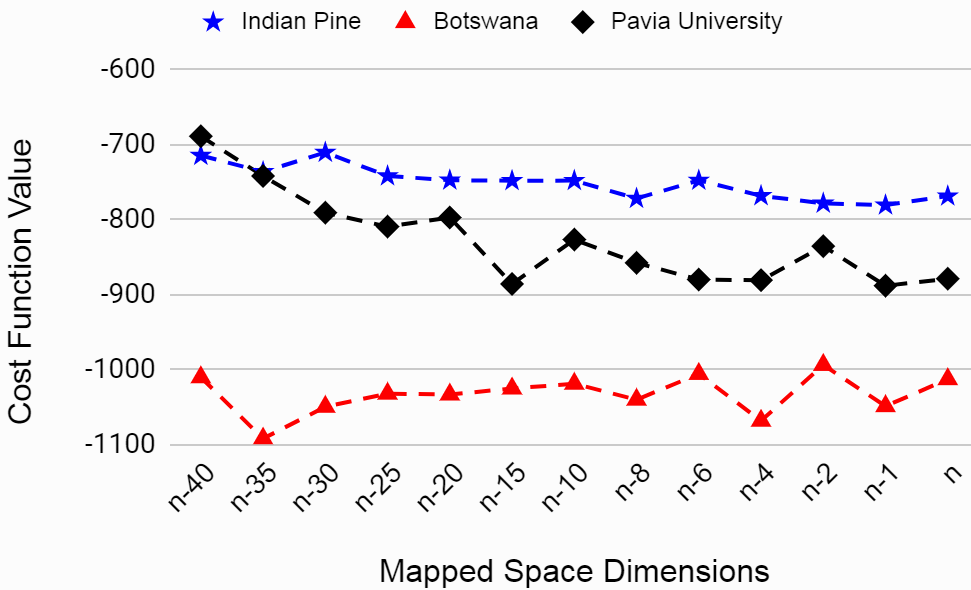}
		%\caption{}%{Botswana dataset}
		%\label{fig:IndP10DR}
		%\end{subfigure}%
		\caption{Comparison of cost function values and mapped space dimension (n: Original dimension of the datasets).}
		\label{fig:DRhsi10}
	\end{figure}
	%#########################################################################################################

	% % Please add the following required packages to your document preamble:
	% % \usepackage{graphicx}
	% \begin{table}[]
	% \centering
	% \caption{My caption}
	% \label{my-label}
	% \resizebox{\textwidth}{!}{%
	% \begin{tabular}{c|cc|cc|cc}
	% \hline
	% Datasets & \multicolumn{2}{c|}{Indian Pines} & \multicolumn{2}{c|}{Botswana} & \multicolumn{2}{c|}{Pavia University} \\ \hline
	% $N_w = N_b$ & Original Space & Projected Space & Original Space & Projected Space & Original Space & Projected Space \\ \hline
	% 3 & 52.86 & 60.11 & 85.96 & 88.01 & 68.90 & 72.27 \\ \hline
	% 5 & 53.50 & 57.82 & 87.65 & 86.91 & 66.30 & 68.82 \\ \hline
	% 7 & 52.75 & 58.57 & 86.44 & 86.88 & 65.98 & 74.32 \\ \hline
	% 9 & 56.43 & 60.53 & 89.12 & 90.39 & 67.30 & 73.93 \\ \hline
	% 11 & 54.87 & 57.53 & 86.06 & 88.18 & 67.27 & 73.19 \\ \hline
	% 13 & 55.94 & 59.44 & 86.70 & 86.14 & 67.08 & 74.05 \\ \hline
	% \end{tabular}%
	% }
	% \end{table}
	%########################################################################################################

	\begin{figure*}[h!]
		\centering
		\begin{subfigure}{.33\textwidth}
			\centering
			\includegraphics[width=.98\linewidth]{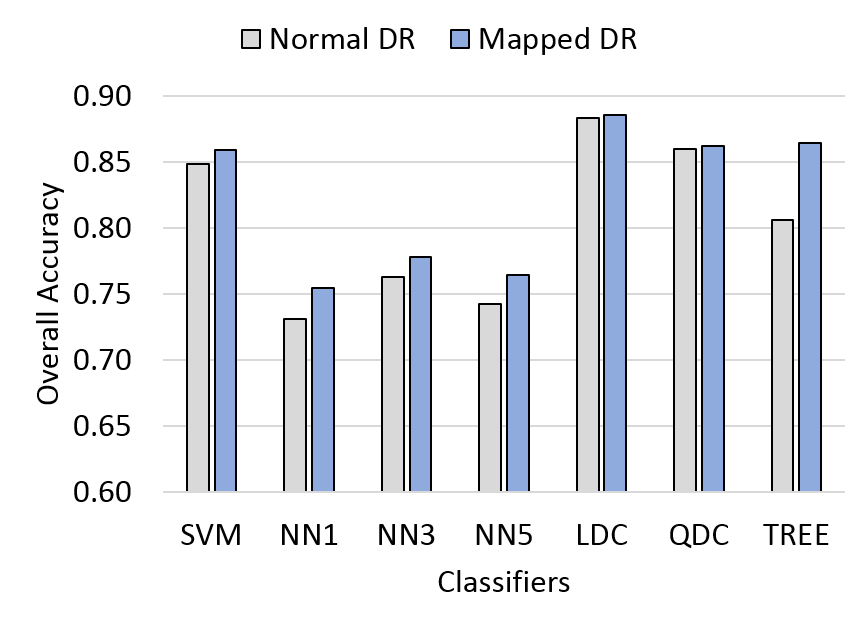}
			\caption{}%{Botswana dataset}
			\label{fig:PCA20DimBot}
		\end{subfigure}%
		\begin{subfigure}{.33\textwidth}
			\centering
			\includegraphics[width=.98\linewidth]{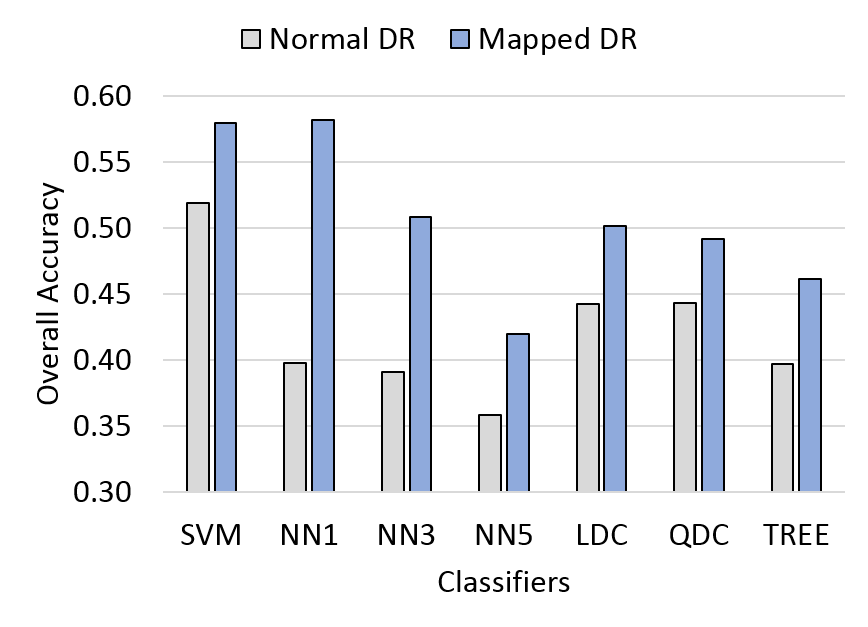}
			\caption{}%Indian-pines dataset}
			\label{fig:PCA20DimIndP}
		\end{subfigure}
		\begin{subfigure}{.33\textwidth}
			\centering
			\includegraphics[width=.98\linewidth]{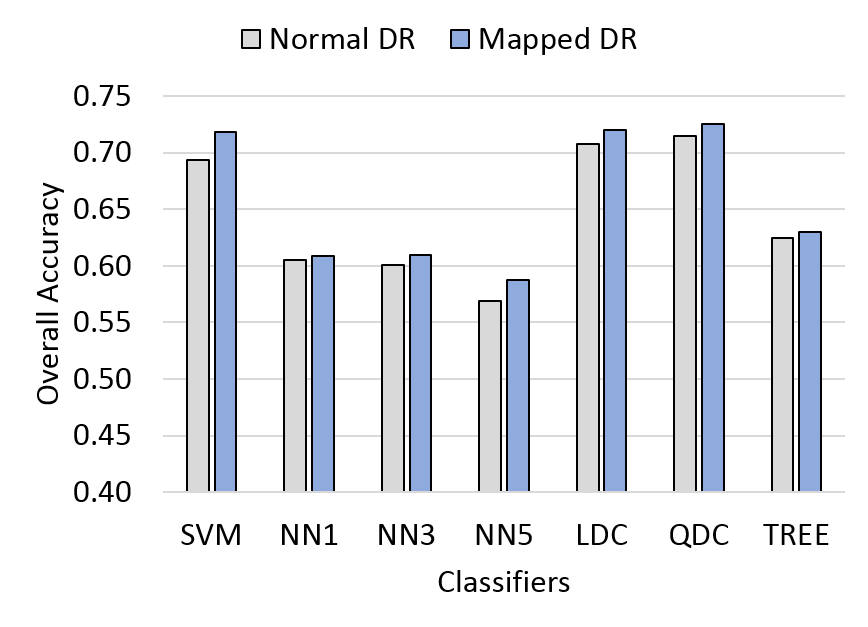}
			\caption{}%Pavia University dataset}
			\label{fig:PCA20DimPavU}
		\end{subfigure}  
		
		\caption{Comparison of classification performances of PCA with various classifiers (SVM, NN1, NN3, NN5, LDC, QDC and Decision Tree) on three different HS datasets (Botswana (\ref{fig:PCA20DimBot}), Indian Pines (\ref{fig:PCA20DimIndP}) and Pavia University (\ref{fig:PCA20DimPavU})) for two spaces (original and mapped discriminative space).}
		\label{fig:PCAhsi10}
	\end{figure*} 
	%################################################################################
	\begin{figure*}[h!]
		\centering
		\begin{subfigure}{.33\textwidth}
			\centering
			\includegraphics[width=.98\linewidth]{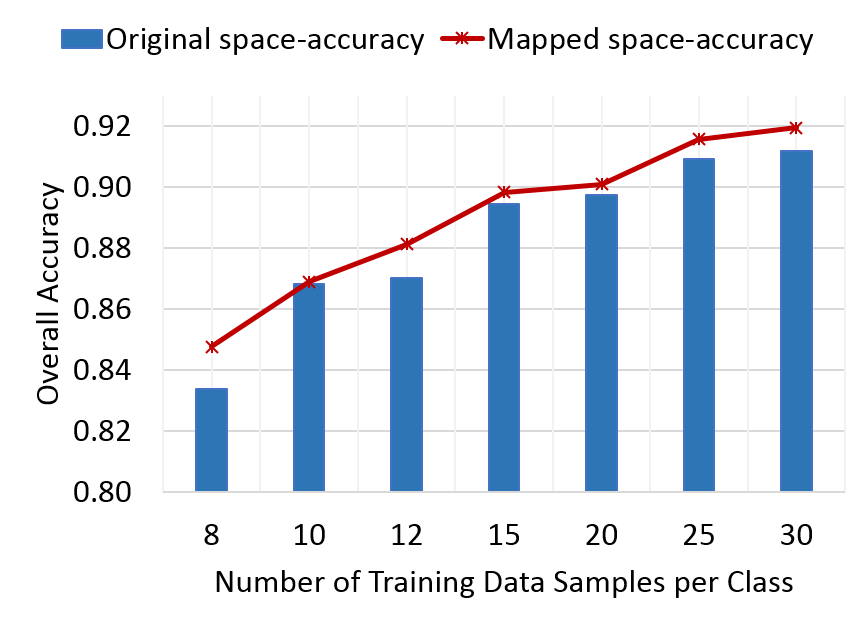}
			\caption{}%{Botswana dataset}
			\label{fig:SVMBots}
		\end{subfigure}%
		\begin{subfigure}{.33\textwidth}
			\centering
			\includegraphics[width=.98\linewidth]{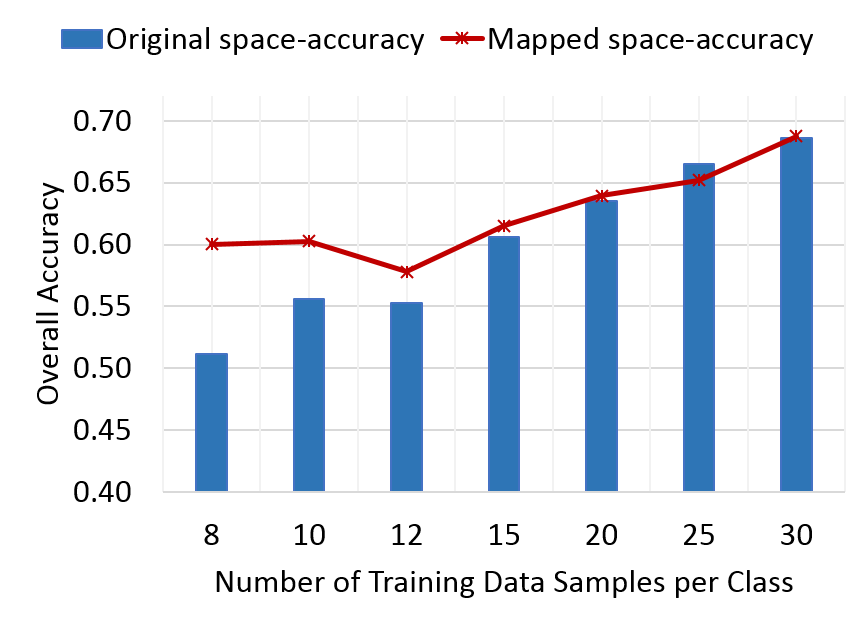}
			\caption{}%Indian-pines dataset}
			\label{fig:SVMIndP}
		\end{subfigure}
		\begin{subfigure}{.33\textwidth}
			\centering
			\includegraphics[width=.98\linewidth]{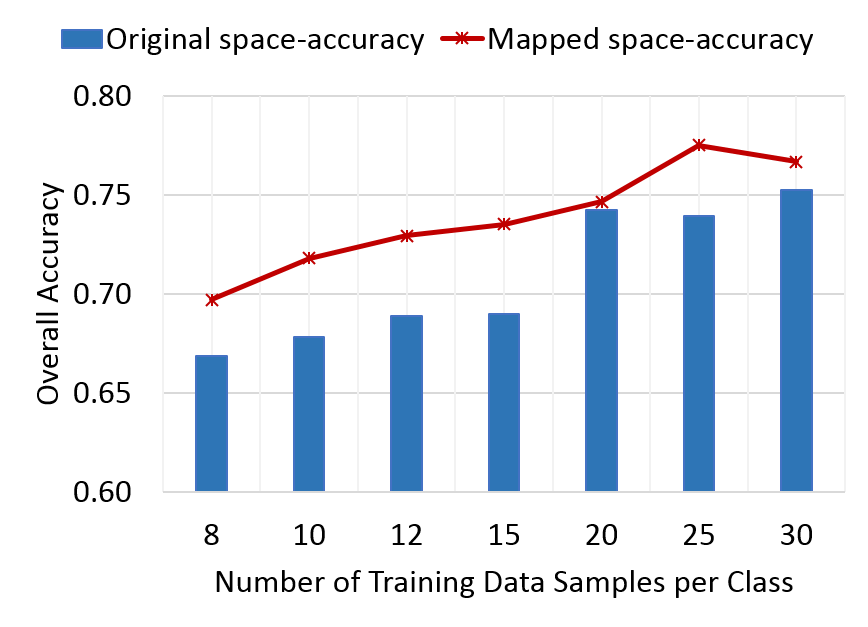}
			\caption{}%Pavia University dataset}
			\label{fig:SVMPavU}
		\end{subfigure}  
		
		\caption{%\textbf{Change figure legends: accuracy on original space-accuracy on mapped space; axis: number of training data samples per class} 
			Comparison of classification accuracy with SVM classifier using different number of training samples per class on three different HS datasets (Botswana (\ref{fig:SVMBots}), Indian Pines (\ref{fig:SVMIndP}) and Pavia University (\ref{fig:SVMPavU})) for two spaces (original and mapped discriminative space).}  
		\label{fig:SVMtrnSampl_3HSI}
	\end{figure*}

	\subsection{Parameters Sensitivity}
	A few samples were randomly selected from each class for training while the rest data were used for testing purpose. Unless specified, we used $10$ random data samples from each class for training. The dimensions chosen for the HS datasets are follows: Indian pine is $199$, Botswana is $144$ and Pavia university is $102$. 
	We employed different types of classifiers to compare the classification accuracy of HS data, such as support vector machine (SVM), K-nearest neighbor (KNN $(K =1,3,5)$), linear discriminant classifier (LDC), quadratic discriminant classifier (QDC), and decision tree (TREE). We used linear kernel for SVM classifier. %The intra and inter-class nearest neighbors $N_w$ and $N_b$ used for constructing the graph were empirically set to $9$.
	
	Table~\ref{tab:knn_OA} provides the statistics of the OA with respect to the variation of number of nearest neighbors ($N_w$ and $N_b$) for three HS datasets while fixing the number of train data samples to $10$ and using SVM as the classifier. In practice, the number of nearest neighbors are always equal for both between-class and within-class ($N_b = N_w$) to avoid the data imbalance. As observed from the Table~\ref{tab:knn_OA}, the SVM classifier gives better accuracy when $N_b = N_w =9$ for two datasets out of three. Hence, we chose the value $N_b = N_w =9$ for subsequent experiments. 
	
	Fig.~\ref{fig:DRhsi10} shows the variations of cost function values with respect to the varied mapped space dimensions. In this experiment, we varied the dimensions of the data in the mapping space from $n$ (original dimension) to $n-40$ and observe the cost function values. In fig.~\ref{fig:DRhsi10} the cost function value is minimum for Indian pine and Pavia university data sets when the mapping dimension is one less than the original data dimensions ($n-1$). In case of Botswana data set, the cost function value is least when mapped space dimension is $n-35$ but it gives competitive results for the $n-1$ mapping dimension. Since, all the three datasets produce consistently better performance (cost function value) on the dimension $n-1$ dimensions in the mapping space, We chose that dimension ($n-1$) for rest of the experiments.
	
	\subsection{Classification Results With Different Classifiers}
	We compared the performances of different classification methods on original and newly mapped space without applying DR techniques.

	Table~\ref{tab:Accuracy_comparison} provides the statistics of the average accuracy (AA), kappa ($\kappa$), and overall classification accuracy (OA) on two spaces (original space and mapped space) for three HS datasets. From Table~\ref{tab:Accuracy_comparison}, we can observe that
	\begin{itemize}
		\item The OA, AA, and $\kappa$ coefficients for all the classifiers on the projected discriminative space always outperforms their counterpart on original data space.
		\item Significant improvements in OA can be observed in Indian Pines and Pavia database on the projected space. 
		\item The performance of SVM is better compared to 1NN or 3NN in the original space. However, the performance of k-NN classifiers are superior to SVM in the projected space. It can be observed that the 1NN outperforms all other classifiers in the projected space and achieves the best performance score. Specifically, in Botswana dataset, it achieves an OA of 96.45\% and an AA of 96.67\%, which is quite remarkable. Similar performance improvements can also be observed in Pavia dataset.
	\end{itemize}
	The results confirm that the learned projection matrix projects the data onto a more discriminative space, which in turn improves the classification accuracies of the HSI data. Similarly, Table~\ref{tab:time} gives the statistics of the training time and testing time for various training sample of Pavia university data set. %This table shows the time for different training samples of Pavia University dataset.%The major time is consumed during training of the data. Once the training is over and the projection matrix is found, the testing time is simply the matrix multiplication operation time. 
	
	\subsection{Classification Results With Different DR Techniques}
	In our experiments, we used four different DR methods such as PCA, LDA, kernel PCA (KPCA) and marginal fisher analysis (MFA) followed by classification using linear SVM. The performance measurements are reported in Fig.~\ref{fig:DRhsi10}. For these experiments, we used $10$ samples from each class for training and the rest as testing samples. The number of reduced dimensions were varied from $20-60$ with a step size of $5$. In addition, we reported the classification performance of other classifiers on the original and projected space after reducing the dimension to 20 using PCA in Fig.~\ref{fig:PCAhsi10}. %Fig.~\ref{fig:SVMtrnSampl_3HSI} shows the experiment results of  
	
	Fig.~\ref{fig:DRhsi10} shows the variations of the overall accuracies of different DR methods with respect to different reduced dimensions on their original data space and newly mapped discriminative space. We can observe that all the DR methods on the proposed discriminative space outperform their counterparts on the original space for various reduced dimensions. In the case of Indian pine data in Fig.~\ref{fig:IndP10DR} all the DR methods in the proposed discriminative space significantly outperforms the DR in original space. Similar performance scores can also be observed in Fig.~\ref{fig:Bow10DR} and \ref{fig:PavU10DR} with a few exceptions. The performance in Fig.~\ref{fig:DRhsi10} proves the robustness of the proposed discriminative space to the variation of number of reduced dimensions in DR methods.
	
	The performance of different classifiers with PCA are shown in Fig.~\ref{fig:PCAhsi10}. In these experiments, we considered multiple classifiers, such as SVM, 1NN, 3NN, 5NN, LDC, QDC and decision tree on original and projected space. %Fig.~\ref{fig:PCAhsi10} shows the robust discriminative property of the projected space when we apply PCA with various classifiers such as SVM, 1NN, 3NN, 5NN, LDC, QDC and decision tree on three HSI datasets. 
	It can be observed in Fig.~\ref{fig:PCA20DimBot} that the overall classification accuracy of PCA in the proposed discriminative space mostly outperforms the accuracy on the original space for Botswana dataset. PCA with classifier LDC and QDC also achieved an equivalent performance for Botswana dataset. In the case of Indian pine dataset in Fig.~\ref{fig:PCA20DimIndP}, the accuracy on the proposed space outperforms the accuracy on the original space by a large margin for all the classifiers. Similarly, the PCA on proposed space also performs better for Pavia university dataset (Fig.~\ref{fig:PCA20DimPavU}).  
	
	\subsection{Classification Results With Different Train Data}
	We also experimented the effect of variation of training data size against the classification accuracy. The experimental results for the performance of the original and projected data space without dimensionality reduction are shown in Fig.~\ref{fig:SVMtrnSampl_3HSI}. As can be seen in Fig.~\ref{fig:SVMBots}, the performance is better in the proposed discriminative space for different training sample size in Botswana dataset. Similar trends are also observed for Indian  Pines (Fig.~\ref{fig:SVMIndP}) and Pavia University (Fig.~\ref{fig:SVMPavU}) dataset. This signifies the effectiveness of the proposed method irrespective of the size of the available data for training.

\section{Conclusion} \label{conclusion}
%\input{conclusion}

%The conclusion goes here.\cite{harandi2017dimensionality}
This work proposed a novel approach in which the data are linearly transferred from their original data space to a more discriminative space to improve the classification performance. The proposed method finds a projection matrix in a supervised manner by solving the nonlinear cost function to preserve the intrinsic geometrical manifold while improving the class discrimination. The experiments show that among the classifiers the 1NN classifier gives upto $8-10 \%$ and among DR methods KPCA with SVM classifier gives upto $5-7 \%$ improved performance in projected space for all the three datasets.  %The experimental results on three popular HSI datasets show that the DR algorithms and classification algorithms achieves significant performance improvement on the proposed discriminative space. 
The promising experimental results of different classifiers with varying number of training data samples on the proposed discriminative space demonstrate its robustness as well as generic applicability.      

% In addition, the results of various classification accuracy of different classifier and different training data samples on the proposed discriminative space shows its robustness.

\bibliographystyle{IEEEtran}
\bibliography{mm_bib}


% \end{IEEEbiography}



\end{document}